\pdfoutput=1

\documentclass[11pt]{article}

\usepackage[]{acl}

\usepackage{times}
\usepackage{latexsym}

\usepackage[T1]{fontenc}

\usepackage[utf8]{inputenc}

\usepackage{microtype}
\usepackage{makecell}

\usepackage{soul}
\usepackage{url}
\usepackage[utf8]{inputenc}
\usepackage{graphicx}
\usepackage{amsmath}
\usepackage{booktabs}
\usepackage{colortbl}
\usepackage{multirow}
\usepackage{float}	
\usepackage{dblfloatfix}
\usepackage{setspace}  

\usepackage{xcolor}	

\definecolor{ForestGreen}{rgb}{0.13, 0.55, 0.13}
\definecolor{skyblue}{HTML}{46C5DD}	

\newcommand{\eat}[1]{}
\newcommand{\red}[1]{\textcolor{red}{#1}}
\newcommand{\blue}[1]{\textcolor{blue}{#1}}
\newcommand{\orange}[1]{\textcolor{orange}{#1}}
\newcommand{\green}[1]{\textcolor{ForestGreen}{#1}}

\newcommand{\camera}[1]{\textcolor{black}{#1}}

\newcommand{\sgap}{\hspace*{2mm}}

\newcommand{\gapxx}{\hspace*{4mm}}

\newcommand{\bfit}[1]{\textbf{\textit{#1}}}

\newcommand{\TeachMe}{TeachMe}
\newcommand{\teachme}{TeachMe}

\usepackage{quoting}
\newenvironment{myquote}{                   
  \parskip 0mm \begin{quoting}[vskip=0mm,leftmargin=2mm]}{
\end{quoting}}
\newenvironment{ite}{                     
     \parskip 0cm \begin{itemize} \parskip 0cm \parsep 0cm \itemsep 0cm \topsep 0cm}{
        \end{itemize}} 
\newenvironment{enu}{                   
     \parskip 0cm \begin{list}{}{\parsep 0cm \itemsep 0cm \topsep 0cm}}{
       \end{list}} 
\newenvironment{des}{                 
     \parskip 0cm \begin{list}{}{\parsep 0cm \itemsep 0cm \topsep 0cm}}{
       \end{list}} 

\newcommand{\quotebox}[1]{\begin{myquote}\fbox{\parbox{0.9\columnwidth}{#1}}\end{myquote}}

\usepackage{algorithm}
\usepackage{algorithm}
\usepackage[noend]{algpseudocode}
\usepackage{ulem}

\title{Towards Teachable Reasoning Systems: Using a Dynamic Memory of
User Feedback for Continual System Improvement}

\author{Bhavana Dalvi Mishra, Oyvind Tafjord, Peter Clark \\
Allen Institute for AI, Seattle, WA \\
{\texttt{\{bhavanad,oyvindt,peterc\}@allenai.org}}
}

\begin{document}
\maketitle

\begin{abstract}
Our goal is a teachable reasoning system for question-answering (QA), where a user can interact with faithful answer explanations, and correct its errors so that the system improves over time. Our approach is to augment a QA model with a dynamic memory of user feedback, containing user-supplied corrections to erroneous model beliefs that users identify during interaction. Retrievals from memory are used as additional context for QA, to help avoid previous mistakes in similar new situations - a novel application of memory-based continuous learning. With simulated feedback, we find that our system (called \TeachMe{}\footnote{\camera{Supplementary data and models are available at    \url{https://allenai.org/data/teachme}}}) continually improves with time, and without model retraining, requiring feedback on only 25\% of training examples to reach within 1\% of the upper-bound (feedback on all examples). Similarly, in experiments with real users, we observe a similar trend, with performance improving by over 15\% on a hidden test set after teaching. This suggests new opportunities for using frozen language models in an interactive setting where users can inspect, debug, and correct the model's beliefs, leading to improved system's performance over time.
\end{abstract}

\section{Introduction}

\begin{figure}[t]
\centering
     \includegraphics[width=1.0\columnwidth]{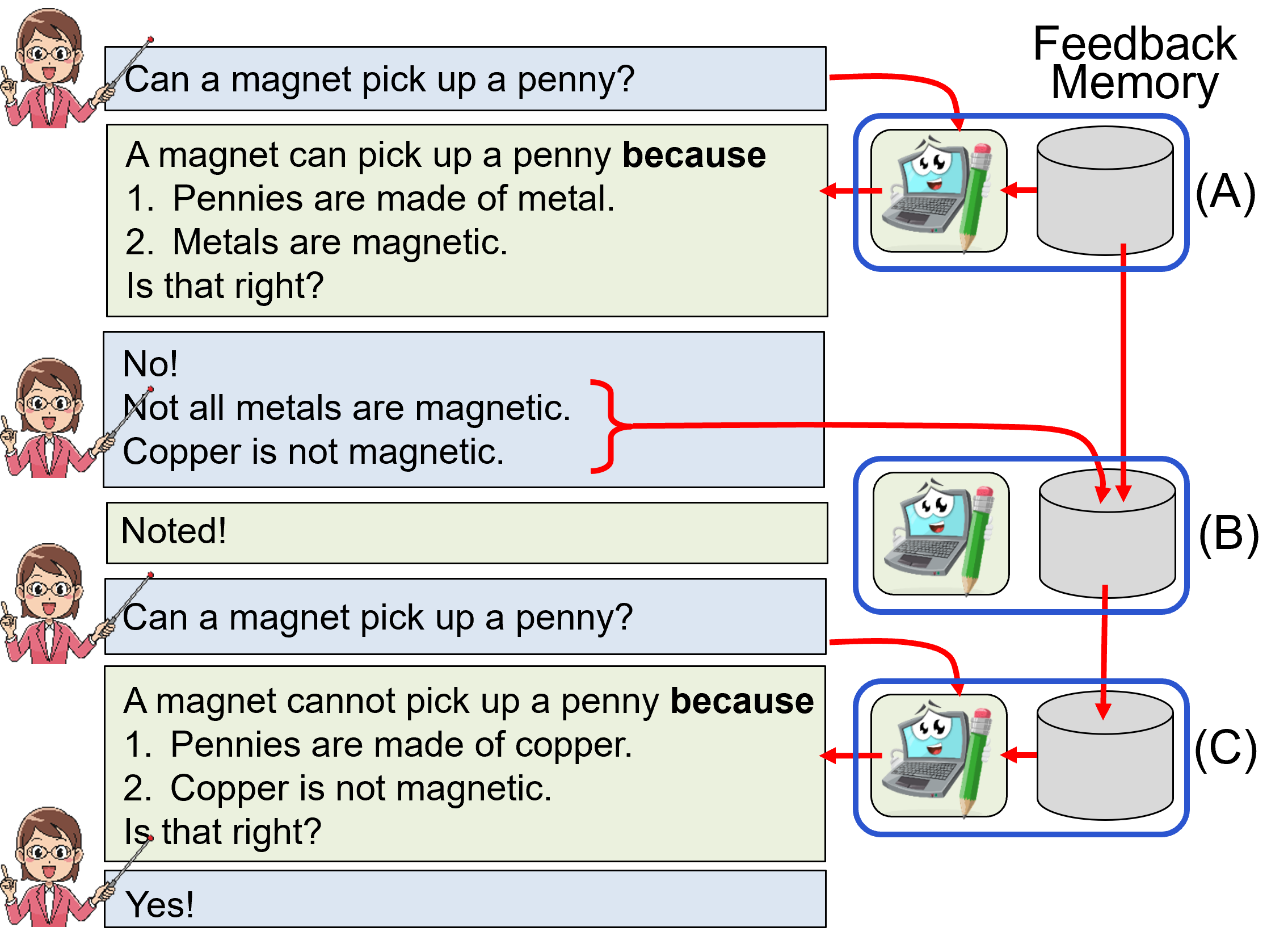}	   
\caption{\TeachMe{} augments the basic question-answering model with a memory of user feedback. (A) Given a new question, facts retrieved from memory are used as additional context for the model, influencing its answers and proofs. (B) If the user disagrees with an answer, they localize the error in the explanation and offer corrective feedback, which is added to memory. (C) These new facts can then be retrieved if the query is re-asked, helping the system avoid repeating mistakes. Note that these also help improve answers on new, similar questions that are asked later, helping the system improve over time.
\label{feedback}}
\vspace{-3mm}
\end{figure}

Our goal is a teachable question-answering (QA) system - one that a user can interact
with to see faithful explanations for its answers, debug errors, and correct them so that the
system gradually improves over time (sometimes referred to 
as explanatory interactive machine learning (XIL) \cite{xil}). 
While the benefits of such a system are evident \cite{Lakkaraju2022RethinkingEA},
the challenges are evident also: despite recent progress in explainability \cite{wiegreffe2021teach},
it is often hard to understand how a model
arrived at an answer, and even harder to correct it if it made a mistake.
In contrast, people are typically able to provide a {\it chain of reasoning} for their decisions,
and may change their mind if a flaw in their knowledge or reasoning is exposed.
Our goal is to similarly have machines provide reasoned answers
to questions, showing how the answer follows from its internal knowledge
(and possibly externally available information), and where it is capable of changing its
answer if errors in that knowledge are identified.

Our approach has three components. First, the system produces answers
supported by an entailment-based chain of reasoning, showing how the answer
follows from the system's own internal beliefs\footnote{
We here adopt a simple definition of belief, namely that a model believes X if it answers "yes" to the
question "Is X true?". Other definitions could also be used.}. 
Second, if an answer is wrong, a user can inspect the reasoning
to diagnose and correct the failure. For example in Figure~\ref{feedback}, the system incorrectly
concludes that ``a magnet can pick up a penny'' from its over-general (false)
belief that ``metals are magnetic''. The user can thus correct the mistake
by asserting that ``not all metals are magnetic'', in particular copper.
Finally, to store and apply the user's feedback, we augment the
model with a dynamic memory. Given a new question (or re-asking an old question), \TeachMe{} retrieves user-supplied facts
from this memory. These are then used as context while generating an
entailment-supported answer to the question, e.g., step (C) in Figure~\ref{feedback}.
This helps override prior, erroneous model beliefs, thus
biasing \TeachMe{} to avoid similar mistakes in future -- a novel 
application of memory-based continual learning to belief
maintenance, in which the model itself remains fixed (frozen)
and retraining is not required.

We evaluate \TeachMe{} using both simulated and real user feedback.
With simulated feedback, using two existing datasets 
OBQA \cite{obqa} and QuaRTz \cite{quartz}, 
we find that \TeachMe{} is able to continuously improve with time,
without retraining, requiring only a quarter of the feedback annotations available
in the original dataset to reach within 1\% of the upper-bound (using all 
gold annotations). Similarly with real users, we find that after they interact
with \TeachMe{} on a small set of questions, the system's performance on
a hidden test set similarly improves (by over 15\%) without retraining.
Our contributions are thus:
\begin{enu}
\item[1.] A novel, memory-augmented architecture enabling user corrections
to help override erroneous model beliefs, thus allowing the overall
system to gradually improve with time, without model retraining (the
runtime model remains frozen).
While memory-based architectures have been used previously, ours is the
first to show that user-provided and model-internal beliefs can be integrated
together for systematic reasoning.
\item[2.] A demonstration of the viability of the approach with both
simulated and real users, showing system improvement on hidden test
questions after users ``taught'' the system on a set of training questions.
\end{enu}

\section{Related Work \label{related-work}}
\noindent
{\bf Guiding Frozen Language Models and Memory:}
Our use of context to modify a
(run-time) frozen model's behavior is similar to retrieval-based QA \cite{Ni2019LearningTA,aristo},
where retrieved context can improve QA performance. In our case, however,
retrieval is from a dynamic memory of user-supplied facts, rather than a static corpus,
the memory serving to expand and override model beliefs.
It also can be seen as a form of prompt engineering \cite{gpt3,Rubin2021LearningTR}, except using relevant
facts rather than few-shot QA examples, and with novelty on the interactive collection and
management of those facts.

\eat{
\TeachMe{}'s memory-based feedback is inspired by the feedback mechanism of BeliefBank \cite{beliefbank},
in which retrieved memories were similarly used as context to improve future QA. \camera{In BeliefBank,
however, memories were previous system answers. It self-reflects on simple facts and does not have mechanisms for explaining its reasoning nor being corrected by a user,
while in \TeachMe{} memories are provided by a user, identified through interaction with system explanations.}
}

\TeachMe{}'s memory-based feedback is inspired by the feedback mechanism of BeliefBank \cite{beliefbank},
in which retrieved memories were similarly used as context to guide future QA. \camera{In BeliefBank,
however, memories were previous system answers, without any mechanism for explaining its reasonong
nor being corrected by a user. In contrast, \TeachMe{}'s memories are provided by a user,
identified through interaction with system explanations.}

\TeachMe{}'s memory is also related to work by Tandon et al., where user feedback memories were used but in different ways,
namely to repair erroneous model outputs via post-processing \cite{learning-to-repair},
or to clarify user intent in GPT3 prompts \cite{prompt-editing}. In contrast, \TeachMe{}'s
feedback contains corrections and elaborations to the model's internal beliefs themselves.

\camera{More generally, while the idea of memory for improved performance is not new, our way of using memory is novel: to the best of our knowledge, TeachMe is the first system that allows a user to find, extend, and correct its reasoning errors, and the memory allows the resulting system to improve over time (continual learning).}

\noindent
{\bf Feedback and Interaction:} Interaction has been successfully used to
learn in interactive recommender systems, e.g., \cite{Kang2019RecommendationAA,Li2021SelfSupervisedBP},
\camera{conversational systems, e.g., BlenderBot \cite{blenderbot3},
knowledge graphs \cite{knowbot}, and procedural tasks \cite{Li2020InteractiveTL}.
Interaction has also been used for data augmentation, by having users identify
model biases and provide additional corrective training examples to reduce
those biases \cite{Kaushik2020LearningTD,lu-etal-2022-rationale}.
In contrast,} our work focuses on learning corrective feedback
in the context of {\it reasoning}. Early AI attempts at having users debug rule-based representations
had limited success, e.g., Teiresias \cite{Davis1977InteractiveTO}, Convince \cite{Kim1987ConvinceAC}.
Our work can be viewed as a modern formulation of this goal, using linguistic expressions
of the knowledge stored latently in a model.


\noindent \textbf{Continual Learning:} Finally, our system performs a kind of continual learning \cite{Parisi2019ContinualLL,Carlson2010TowardAA}, aiming to correct specific errors that appear. Recent work has explored ``model editing'' - editing model parameters to fix incorrect answers or add new knowledge \citep{Mitchell2021FastME,de-cao-etal-2021-editing, hase2021beleifs}. However, to date these approaches have only been demonstrated in a limited context (e.g., correcting a single error), and even then can lead to uncontrollable out-of-scope changes \cite{Mitchell2021FastME}. In contrast, our goal is not just to correct a specific error, but to have that correction generalize to new problems, and without damaging the model's basic problem-solving acumen. Thus, our work leaves the model fixed, and seeks improvement in the broader system in which the model is embedded, exploring an alternative and potentially more interpretable architecture towards this goal.

\section{Approach}

We adopt a question-centric approach to teaching and interaction, in which the user (teacher)
asks the system (student) a question that they know the answer to, to probe the system's knowledge.
The system then answers it along with a {\it faithful} entailment-based explanation. If the
system's answer is wrong, the user can interact with the explanation to identify the erroneous system beliefs
that lead to the incorrect answer, and correct them.
Corrections are stored in a dynamic memory used to influence, and ideally improve, future
system behavior.


We instantiate this approach in a system called \TeachMe{}, which has three key components:
\begin{enu}
\item[{\bf 1. Answering Questions:}] Given a user's question, \TeachMe{} searches for an entailment-based
	line of reasoning for different candidate answers, and selects the best. 
\item[{\bf 2. Interaction:}] The user can inspect, locate, and correct errors in the system beliefs
           that led to incorrect answers.
\item[{\bf 3. Dynamic Memory:}] \TeachMe{} maintains a dynamic memory
           of user-corrected beliefs, used to help answer future questions. 
\end{enu}
We now describe each in turn.

\subsection{Answering Questions \label{answering}}

The key requirement of this component is to show how an
answer systematically follows from the model's own beliefs - in other words, provide
an explanation that is both {\it truthful} (reflects the system's own beliefs) and
{\it faithful} (the answer choice follows from those beliefs). Beyond this,
\TeachMe{} is agnostic as to how this is done - we describe our approach below,
but others could be used.

\eat{
While not the focus of this paper, we briefly describe how \TeachMe{} generates an answer
to a user's question, plus an explanation of how that answer is entailed by facts
that the system believes are true. These provide a basis for subsequent interaction
with the user.}

\subsubsection{Candidate Hypothesis Generation \label{qa2d}}

Given a question from the user, \TeachMe{} first generates candidate answers and converts
these into declarative hypotheses (e.g., ``Is the sky (A) blue (B) yellow'' $\rightarrow$
\{ $H_1$ = ``The sky is blue.'', $H_2$ = ``The sky is yellow.'').\footnote{Conversion of a QA pair to a declarative hypothesis D uses a custom T5-11B model trained on the QA2D dataset \cite{qa2d}.}
An $N$-way multiple choice question yields $N$ hypotheses.
A true/false question yields 2 hypotheses. For open-ended questions, \TeachMe{}
first collects $N$ candidate answers generated by an external QA system (we use Macaw \cite{macaw})
using nucleus sampling,
then forms $N$ hypotheses from them.

\begin{figure}
\centering
     \includegraphics[width=1\columnwidth]{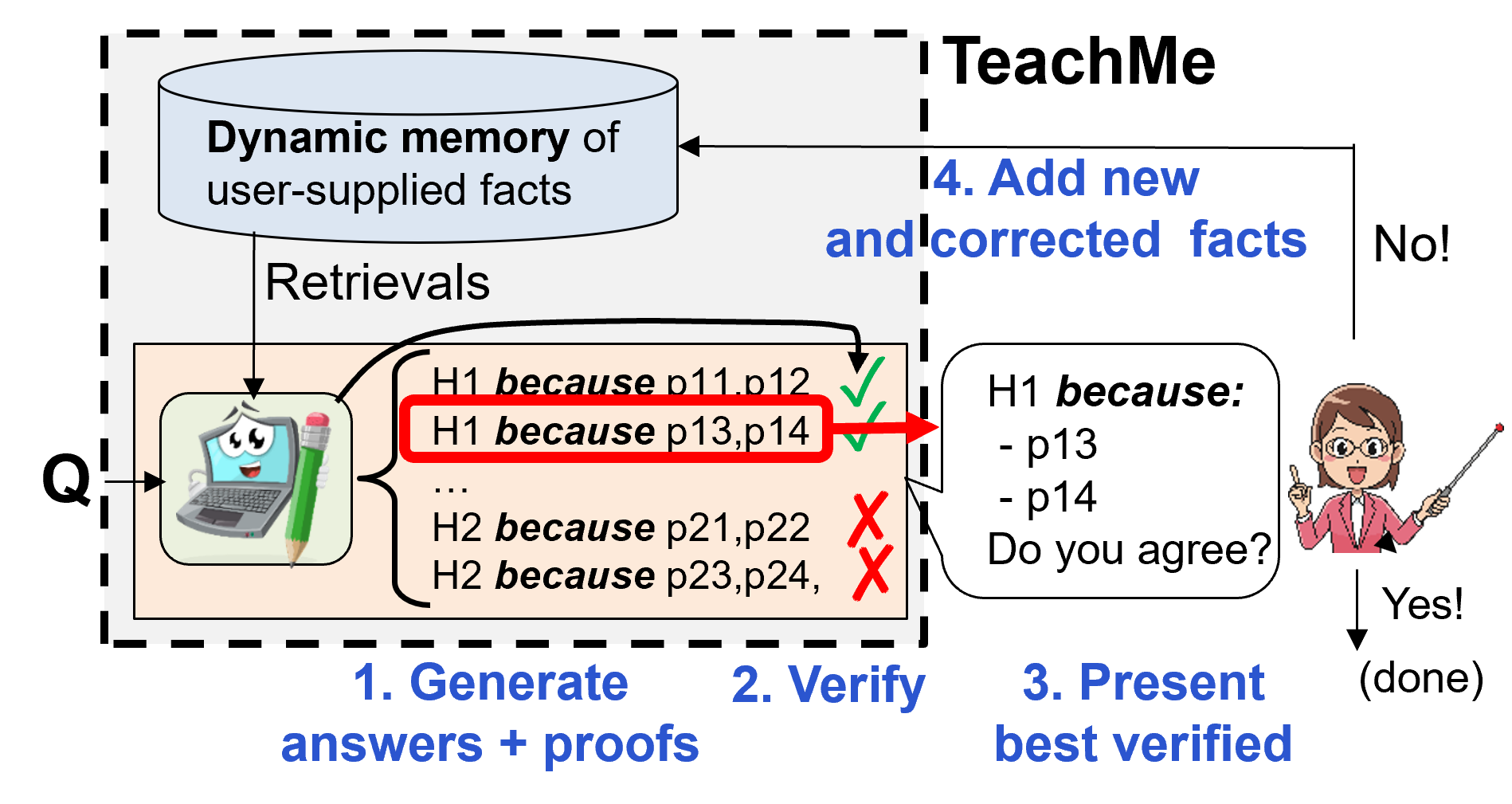}	   
\caption{\TeachMe{}'s architecture contains a model and memory. Given a question,
\TeachMe{} generates multiple answers and proofs, discards those not consistent with its own beliefs (verification),
and presents the best to the user (teacher). If the answer is wrong, the user interacts to identify erroneous
model beliefs, and add corrections to memory, which in turn modifies future QA behavior without model retraining.
\label{architecture}}
\vspace{-2mm}
\end{figure}

\subsubsection{Entailment Proof Generation \label{generating-entailment-trees}}

\TeachMe{} then tries to generate a ``proof''\footnote{
We use the word ``proof'' for convenience but note that the term is somewhat approximate,
as entailment ``proofs'' do not have the guarantees of formal, deductive proofs.}
for each hypothesis $H$, where here a proof means a set of premises (sentences)
such that the hypothesis clearly follows from (is entailed by) the premises.

\begin{figure*}[!t]
\centering{
{\includegraphics[width=1\textwidth]{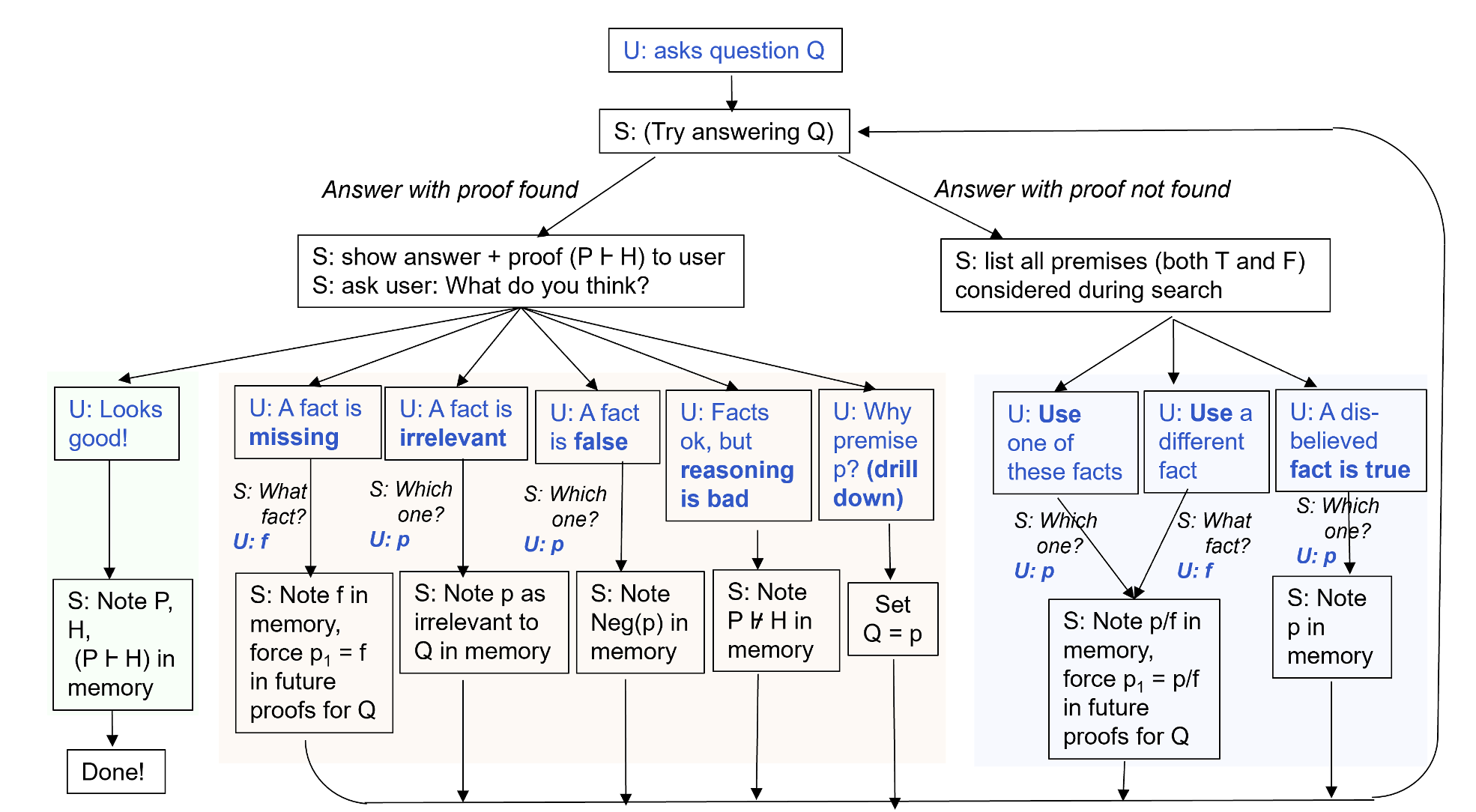}}
}
\caption{\camera{TeachMe's dialog tree, showing the different ways a user can interact with the system.}} \label{fig:interaction_flow}
\end{figure*}

\camera{There are several ways such a proof might be generated. In our case we use Entailer\footnote{\camera{Entailer models are available at \url{https://allenai.org/data/teachme}}} \cite{entailer}, a T5-11B model trained on EntailmentBank -- a large, existing dataset of such textual entailment proofs \cite{entailmentbank}.
} The input to the model is a hypothesis $H$,
plus optionally the question $Q$, answer $A$, and a context of relevant sentences $C$,
and the output is $P$, a set of premises (sentences) that entail $H$.
To ensure the proof is {\it truthful}, the system asks itself ``Is $p_i$ true?'' for each
premise $p_i$, reflecting our definition of belief (footnote 1), and if not,
the proof is rejected. Finally, the proofs are scored, and the final answer
is the hypothesis with the highest-scoring proof (hence the answer is
{\it faithful} to the proof). An example result ($H$ \bfit{because} $P$) is: \vspace{1mm}\\
\fbox{\parbox{0.97\columnwidth}{\small{
Plants require CO2 to make their own food \bfit{because:} \\
1. a plant requires CO2 for photosynthesis \\
2. Plants create food through photosynthesis}}} 
Full details are given in \cite{entailer}.

\camera{
Note that such proofs could be generated in other ways also, for example using
chain-of-thought style, zero-shot prompting to a large model such as GPT3 \cite{Wei2022ChainOT} (continuation in gray):\\
\fbox{\parbox{0.97\columnwidth}{\small{
Plants require CO2 to make their own food. \\
Explain the last statement with a 2-step reasoning chain: \\
1. \sethlcolor{lightgray}\hl{Plants use photosynthesis to produce their own food.  \\
2. Photosynthesis requires CO2 in order to create glucose from water and sunlight.}
}}}
followed by verification steps to ensure that each premise and the entailment itself
were believed by the model, i.e., reflected the model's ``beliefs'' about the world,
and to score them. Again, these could be performed using zero- or few-shot prompting.}


\eat{
We also desire that the premises express the model's own beliefs about the world
(i.e., the proof is {\it truthful}) so any user feedback is a meaningful reflection
about the system's beliefs. To do this, we use self-querying, i.e., the system asks
itself ``Is $p_i$ true?'' for each premise $p_i$, reflecting our definition of
belief (footnote 1). It also asks itself if the entailment as a whole appears valid.
Again, we include these two functionalities when training the model, and use
the model for these tasks. Training details are given in Appendix~A.

Finally the hypothesis with the highest-ranked\footnote{When self-querying,
we obtain a score (confidence) that each belief $p_i$ is true, and a score
that the entailment itself is true. For ranking purposes, we take the overall proof score
as the product of all these, i.e., approximating the scores as probabilities.}
}

\subsection{Interaction \label{interaction}}



Given the system's answer plus entailment proof, users can interact with the system
to inspect, debug, and correct system mistakes via a simple user interface.
Specifically, if an answer is entailed by the system's beliefs, and the answer
is wrong, then either one of those beliefs must be wrong or the entailment itself
must be invalid. To repair this, the user can correct an erroneous model belief that
they identify (or block the entailment itself, by adding it to a list of blocked entailments).
Corrected model beliefs are stored in the dynamic memory. For future questions, retrievals from
memory are used as question context to bias the model away from the prior erroneous beliefs. In this way, user-corrected
beliefs may also help on new, unseen questions. 

\camera{Figure \ref{fig:interaction_flow} shows the overall flow of interaction in \TeachMe{}.
The user first poses a question, then the system attempts to answer it with
a supporting chain of reasoning. If found, the user can critique it and (if
wrong), correct the system's misunderstandings and/or lack of knowledge
using the actions shown. If not found, the system tries to elicit user
guidance by showing the user the facts it considered during search for an
answer, to which the user can respond.
After making a change, the system re-attempts the question, and shows
its updated answer and reasoning. When the user is satisfied,
the premises and answer are then noted in memory, for consistency
if those premises are queried about in future.}

A brief example is below, transcribed into textual format  (\camera{the original interactions
are implemented graphically using hyperlinks and text boxes, illustrated in Appendices B and C}).
The user first asks a question, and
the system answers incorrectly. The user volunteers \blue{A penny is made of copper},
and the system retries, but now reveals its erroneous belief that magnets can attract copper.
The user points out this is false, and finally the system succeeds:

\noindent
\fbox{\parbox{0.97\columnwidth}{\small{
\begin{des}
\item[{\bf U:}] \blue{Can a magnet attract a penny?} 
\item[{\bf S:}] A magnet can attract a penny \bfit{because:} \\
   1. A magnet can attract magnetic metals \\
   2. A penny is made of magnetic metal \\
   Do you agree?
\item[{\bf U:}] \blue{No. A penny is made of copper.} \hfill \bfit{[A fact is missing *]} 
\item[{\bf S:}] A magnet can attract a penny \bfit{because:} \\
   1. A penny is made of copper. \\
   2. A magnet can attract copper. \\
   Do you agree?
\item[{\bf U:}] \blue{No. 2 is false.} \hfill \bfit{[A fact is false *]} 
\item[{\bf S:}] A magnet cannot attract a penny \bfit{because:} \\
   1. A penny is made of copper. \\
   2. A magnet cannot attract copper. \\
   Do you agree?
\item[{\bf U:}] \blue{yes}					\hfill		\bfit{[Looks good]}
\end{des}}}}

\normalsize
\noindent
As a result of this interaction, the user-asserted facts (*) have been added to memory:

\begin{center}
\fbox{\parbox{0.6\columnwidth}{
\small{\underline{\bf Memory} \\
{\it \blue{... earlier assertions ...}} \\
\blue{A penny is made of copper.} \\
\blue{A magnet cannot attract copper.}}}}
\end{center}

\normalsize \noindent
Now given a new question, retrievals from memory are added to the context
when answering. Here, the retrievals include ``A magnet cannot attract copper'',
helping the model answer correctly:

\noindent
\fbox{\parbox{0.97\columnwidth}{\small{
\begin{des}
\item[{\bf U:}] \blue{Can a magnet attract a copper pan?} 
\item[{\bf S:}] A magnet cannot attract a copper pan \bfit{because:} \\
   1. A cooper pan is made of copper \\
   2. A magnet cannot attract copper \\
   Do you agree? 
\item[{\bf U:}] \blue{yes}					\hfill		\bfit{[Looks good]}
\end{des}}}}

\noindent
\normalsize
Given its simple nature, we do not make any claims for this interface
in this paper. Rather, it is just one means by which the memory
of corrected model beliefs can be constructed.
Further transcribed examples (from real users) are in Appendix~D.

\subsection{Dynamic Memory \label{sec:model_memory}}

The third component of \TeachMe{} is a dynamic memory,
containing a list of assertions (English sentences), collected through
interaction. The memory serves as a set of additions
and overrides to the model's latent beliefs, and to our knowledge
is the first to show that user-provided and model-internal beliefs can be
integrated together for systematic reasoning.

Given a question, \TeachMe{} retrieves up to $r$ (= 5) sentences from memory using the
question as the search query, using a standard BM25 search algorithm\footnote{We explore
alternative retrieval strategies in Section~\ref{retrieval} later}.
The retrievals are then used as follows:

\paragraph{As Context:} During generation of an answer + proof (Section~\ref{answering}),
retrieved facts are provided as context to the model.
This encourages (but does not force) \TeachMe{} to use these
facts in a generated proof and avoid conflicting facts.
In this way, these user-supplied facts help \TeachMe{} avoid
mistakes that it previously made.

\eat{
using the optional context field (angle) C in its input (Section~\ref{generating-entailment-trees})
to influence its behavior. In our case, as these retrievals are high-quality,
user-approved facts rather than noisy retrievals from a heterogenous
corpus, we train \TeachMe{} to pay more attention to them than if they had
come (say) from the Web (Appendix~A).
This encourages (but does not force) \TeachMe{} to use these
facts in a generated proof.}

\paragraph{Forced Generation:} Given $r$ retrieved sentences,
we also {\it force} \TeachMe{} to explore proofs that use them, to
ensure user-supplied sentences are fully considered by the model.
This is done using forced generation during decoding time, so that
each proof starts with a different sentence as its first premise.
Given $r$ sentences, we generate $r$ forced proofs in this way, plus
a $r+1$ proof without forced generation. This forcing can also be seen
as a way of encouraging diversity in the generations.
Note that many of these proofs may later be rejected if verification fails.
The highest-scoring proof is then selected.
The full algorithm is in Algorithm~\ref{algo:EW_MC_prediction}.

\algrenewcommand\algorithmicindent{1.0em}%
\algrenewcommand\algorithmiccomment[1]{\hfill {\it // #1}}

\begin{algorithm}[t]
\caption{\TeachMe{}'s Overall Control Algorithm \label{algo:EW_MC_prediction}}
\begin{algorithmic}[1]
\small
\Procedure{answer}{$Q$: question, $A$: Answer choices, $M$: memory of useful facts }       
    \State E = $\phi$  \Comment{Initialize proofs so far}
    \State $C$ = search(corpus=$M$, query=concat($Q$,$A$)
    \For{\texttt{ $A_i \in A$}} 
        \State {\it // Generate a hypothesis $H_i$ for each choice $A_{i}$}
        \State $H_i$ = Hypothesis$_i$ = QA2D($Q$,$A_i$) 
        \For{\texttt{ $C_j \in C \cup \{$none$\}$}}  \Comment{for each sent $C_{j}$}
        \State {\it // 1. Generate a proof}
  	\State {\bf generate} a proof $(P_i \vdash H_i)$ of $H_i$ with first
        \State \gapxx premise = $C_j$ (details in Section~\ref{generating-entailment-trees})$^{\dagger}$
        \State {\it // 2. Add the proof to the list of proofs so far:}
        \State E = E $\cup$ <($P_{i}\vdash{}H_i$), $s(H_i)$, $A_i$>
      \EndFor
    \EndFor
    \State <$E_{best},score_{best},A_{best}$> = Max($E$)
    \State {\bf return} answer=$A_{best}$, explanation=$E_{best}$
\EndProcedure
\end{algorithmic}
\footnotesize{
  $^{\dagger}$When the model generates premises $P_i$, the $Q$, $A_{i}$, and $C$ are provided as additional
  model inputs, and the output is constrained to start with $C_j$ (forced generation).} 
\end{algorithm}

\eat{
\TeachMe{} generates $r+1$ alternative proofs,
where each of the first $r$ proofs uses one of the retrieved sentences
as the first sentence in the proof.

This is done using forced generation during decoding time so that the
proof starts with the $r$th retrieved
sentence as its first premise. The last $r+1$ proof is generated
as normal (no forced generation). This forcing can be seen as a way
of encouraging diversity in the generations, as each of the $r$ proofs
necessarily starts with a different premise.

The overall control algorithm is shown in Algorithm~\ref{algo:EW_MC_prediction}.
Given a question Q with answer options A, \TeachMe{} first retrieves potentially relevant sentences $C$ from
memory (line 3). Then, for each answer option, it creates a hypothesis (line 6),
then generates proofs using each sentence in $C$ in turn as
the first premise, as well as free-form generation (line 9).
Finally, the answer option associated with the highest scored proof is returned
(line 14).}

\section{Experiments and Results \label{experiments}}

Our goal is that \TeachMe{}'s memory-augmented architecture will
allow users to teach the system in a {\it general} way, adding to and correcting
model beliefs so that its performance improves on new, unseen questions.
To evaluate this, we use both both simulated and real users. In both
cases, users first provide feedback on a set of training questions,
populating the memory. Then, with no further interaction, we measure whether \TeachMe{}'s
performance has improved on a set of hidden test questions. In all cases, \TeachMe{}'s
model is frozen - any improvements are purely via memory updates.

\subsection{Datasets}

We evaluate with two existing multiple-choice datasets,
OBQA \cite{obqa} and QuaRTz \cite{quartz}. These datasets contain
questions that (typically) require multihop reasoning, along
with a (crowdworker created) gold 1-step entailment proof for every correct answer
option. In addition, among the premises in those gold proofs, one has been
tagged as the ``core'' (most important) fact of the proof (e.g., ``Metals
conduct electricity''), with several questions sharing the  core fact.
These core facts can help us simulate the user feedback.

For meaningful feedback experiments, there should be at least topical overlap
between train (teaching) and test (evaluation) partitions. In OBQA,
this topical overlap occurs naturally because the train/test partitions
were created randomly, meaning that questions based on the same core fact
are distributed between train and test.\footnote{
Note that the questions based on the same core fact are still substantially
different, e.g., there are many questions one can create based on the 
core fact that ``Magnets attract iron.''} QuaRTz, however, was originally
partitioned to remove topical (core fact) overlap between train and test.
As a result, we use just the training partition of QuaRTz, and repartition
it randomly into Train'/Dev'/Test', leading to a natural topical overlap
between the new partitions.

The sizes of the partitions we use are OBQA train/test = 4957/500 examples,
and QuaRTz Train'/Test' = 1348/557 examples.


\subsection{Experiments with a Simulated User \label{simulation}}

We first measure \TeachMe{}'s ability to learn through interaction with
a simulated user (teacher). In this scenario, we consider the teacher working through
the training questions, and behaving as follows: \vspace{1mm} \\
{\bf 1. If \TeachMe{} answers the question correctly} then no action is
taken. This makes the simplifying assumption that the generated chain of reasoning
is also correct. \vspace{1mm} \\
{\bf 2. If \TeachMe{} answers the question incorrectly} then the user will
provide feedback to help correct the system. In the simulated scenario, we
take the core fact in the gold entailment proof as that user feedback: As the
system was wrong, we here assume that either the model did not
know this core fact, or failed to attend to it when trying to generate
a chain of reasoning for the correct answer. The (simulated) user thus aims
to correct this by providing that fact. This new fact is then added to the system's memory,
where it may be recalled and used for future questions to avoid a similar
mistake in future. Although only an approximation, it allows us to assess
whether this failure-driven feedback
also helps on future, unseen questions.

\vspace{1mm}
Once simulated teaching is completed, we then test the system on a hidden 
test set (no further interaction), measuring QA accuracy.

\subsubsection{Configurations}

\noindent
We compare the following configurations, all using the frozen model, i.e.,
evaluating the impact of feedback that a deployed system would receive: \\
{\bf 1. Direct QA (non-teachable):} We measure the model's basic ability to directly
answer the test questions, without using a reasoning chain, using the $H \rightarrow S_d$ angle.
One can loosely think of this as the ``fast thinking'' answer. \vspace{1mm} \\  
{\bf 2. \TeachMe{} (before teaching):} Here we measure \TeachMe{}'s ability
to answer the test questions by generating, scoring, and comparing
entailment proofs for each answer option, when the memory is in its initial state (empty). One can loosely think of this as the ``slow thinking''
answer. \vspace{1mm} \\
{\bf 3. \TeachMe{} (after teaching):} This is at the end of simulated teaching scenario,
after the simulated user provided feedback (the appropriate core fact) for all training
questions that \TeachMe{} answered incorrectly, thus populating the memory. \\ 
{\bf 4. \TeachMe{} ($\approx$ upper bound: feedback for {\it all} answers):} As
an upper bound, we imagine the user providing feedback on {\it all} training questions,
regardless of whether \TeachMe{} answered them correctly. 
To simulate
this, \TeachMe{}'s memory is set to all the core facts used in all training questions.
In this upper-bound scenario, the simulated user is doing approximately the same work as it
took to create the training dataset proofs in the first place.


\begin{figure}
\centering
     \includegraphics[width=0.9\columnwidth]{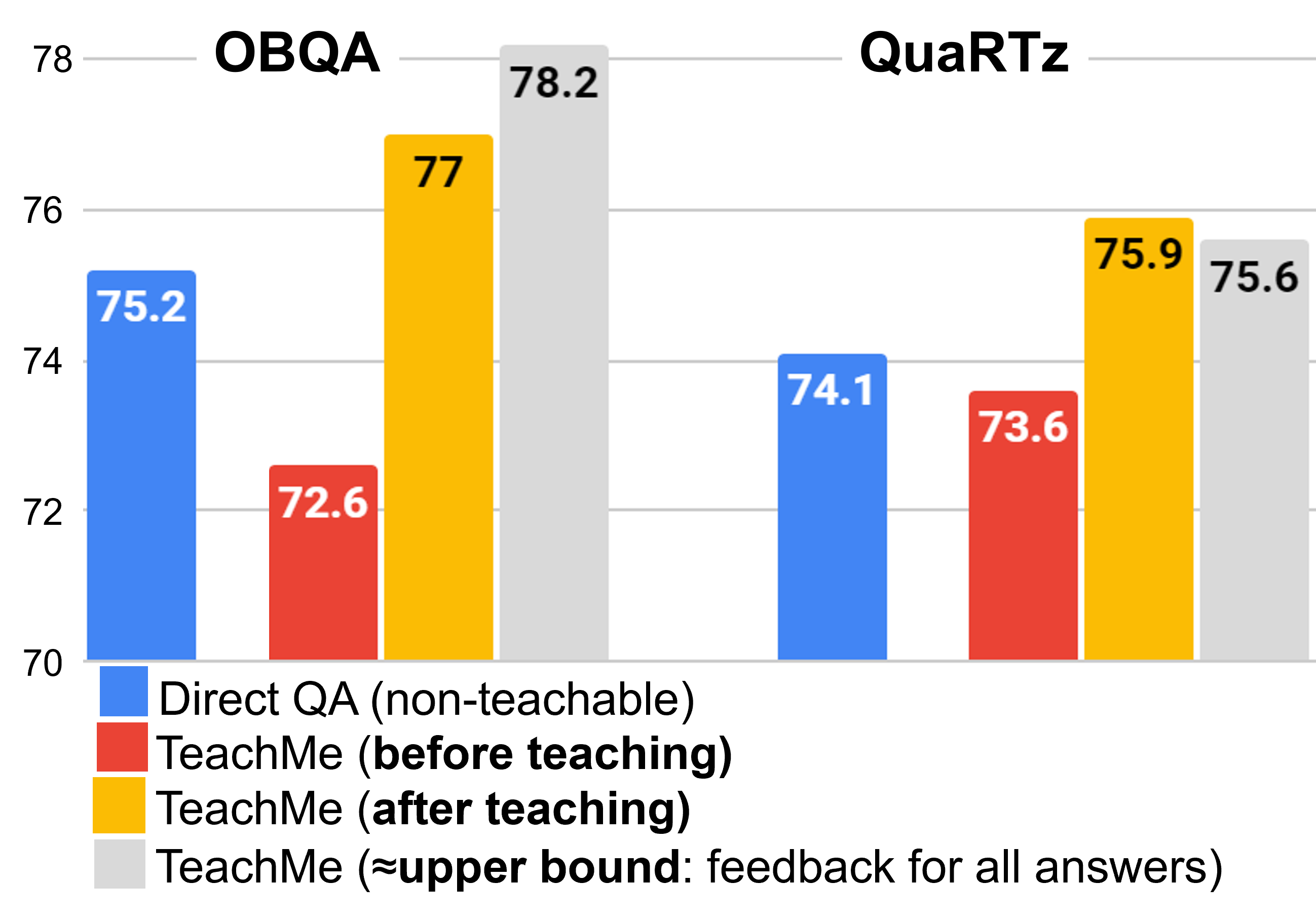}	   
\caption{\TeachMe{}'s performance on the hidden test sets improves with simulated user feedback
(from red to yellow), improving over direct QA and coming close (within $\approx$ 1\%) of the upper bound of
using feedback on all answers (grey). \label{results}}
\vspace{-2mm}
\end{figure}

\subsubsection{Results}

The results are shown in Figure~\ref{results}. Our main findings are as follows: \vspace{1mm} \\
{\bf \TeachMe{}'s Basic Accuracy is Close to that of Direct Answering:}
Comparing \TeachMe{} (before teaching) with direct QA, we see \TeachMe{}'s
proof-based answer accuracy is close, but not quite as good as, the accuracy for direct QA (72.6\% vs. 75.2\% OBQA,
73.6\% vs. 74.1\% QuaRTz). It is encouraging that the scores are loosely comparable,
as it suggests users are critiquing proofs of reasonable quality. A primary cause of
failure is errors by the two verifiers, in particular the entailment verifier
$PH \rightarrow S_e$ sometimes mis-recognizes a bad entailment as valid. 
\vspace{1mm} \\
{\bf Feedback helps on new questions.}
Most significantly, feedback on the training questions has helped improve performance
on the test questions {\bf without requiring model retraining} (OBQA: 72.6\% to 77.0\%;
QuaRTz: 73.6\% to 75.9\%), indicating the viability of the paradigm we are exploring.
The with-memory scores also {\bf exceed the direct QA scores} on both datasets.
\vspace{1mm} \\
{\bf Feedback reaches within 1\% of the upper bound} while only requiring
feedback on $\approx$30\% of the training questions (namely those that the model
answered incorrectly). This suggests that targeted feedback is sufficient
to obtain near-optimal performance, avoiding the high cost of exhaustively
annotating the proofs for all the training questions, as was done in the original
datasets.

\subsubsection{Retrieval Strategies \label{retrieval}}

Facts in memory are indexed by the words in those facts. We also evaluated alternative
indexing strategies, e.g., indexing a fact by the question(s) that used it in
the answer proof, or a combination of question plus fact, but these did not work as well.
Details and results are in Appendix A.


\subsubsection{Improvement with Time}

\begin{figure}
\centering
     \includegraphics[width=\columnwidth]{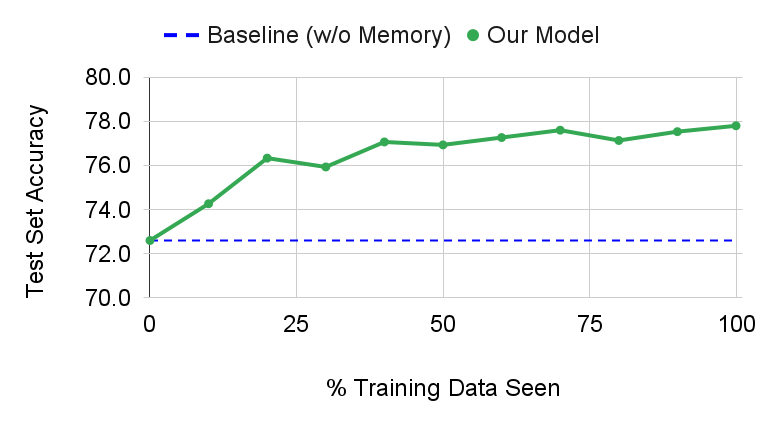}	   
\caption{\TeachMe{}'s performance on OBQA test improves as it sees a larger fraction of training data and stores feedback for wrong answers in its memory.  \label{improvement-with-time}}
\vspace{-2mm}
\end{figure} 

How does \TeachMe{}'s performance improve with time?
To track this, we re-used the OBQA dataset and measured \TeachMe{}'s
performance on the test set as it sees a larger fraction of training data,
storing the feedback for wrong answers it has seen so far in its memory.
The results were averaged over 3 random orderings of OBQA training data,
and are shown in Figure~\ref{improvement-with-time}.
As can be seen, the performance gradually improves as more feedback is collected on failing training questions.
Note that a larger memory does not guarantee better performance, e.g. when training data increases from 20\% to 30\% in Figure~\ref{improvement-with-time}, because
\TeachMe{} may retrieve distracting facts from memory, resulting in spurious proofs supporting wrong answers.

\subsubsection{Analysis}

\subsubsection{Success Analysis \label{success}}

When \TeachMe{} changed its (test set) answer from a wrong answer option (no feedback) to the
correct answer option (with feedback), was that change for a good reason?
Our interest here is
whether \TeachMe{} did indeed recall and use relevant domain knowledge appropriately.
To explore this, we analyzed a random sample of 50 of the 74/500 test cases where such positive flips occurred.
Of these, we found approximately 3/4 resulted from good reasoning, while approximately 1/4 were not.
Comparing the generated and gold test set proofs, we found four groupings,
illustrated in Figure~\ref{successes} and described below (Table~E1 in Appendix~E provides examples of all four): 
   \textbf{28\% (14/50) : the gold core fact} was included in the best scoring proof. 
   \textbf{28\% (14/50) : a relevant core fact} (though not exactly the gold core fact) was used. 
   \textbf{20\% (10/50) : a remotely related fact} was retrieved and used by the model as the first premise in the
   proof due to forced generation (Section~\ref{sec:model_memory}). 
   \textbf{24\% (12/50) : a spurious fact} was retrieved due to word overlap with the question, then the
   model produced an incoherent proof connecting it to the correct answer hypothesis, and
   scored this proof highest. Although this error was advantageous in these cases,
   there are analogous failure cases where a spurious fact changes a previously correct answer to incorrect.

\begin{figure}
\centering
     \includegraphics[width=0.9\columnwidth]{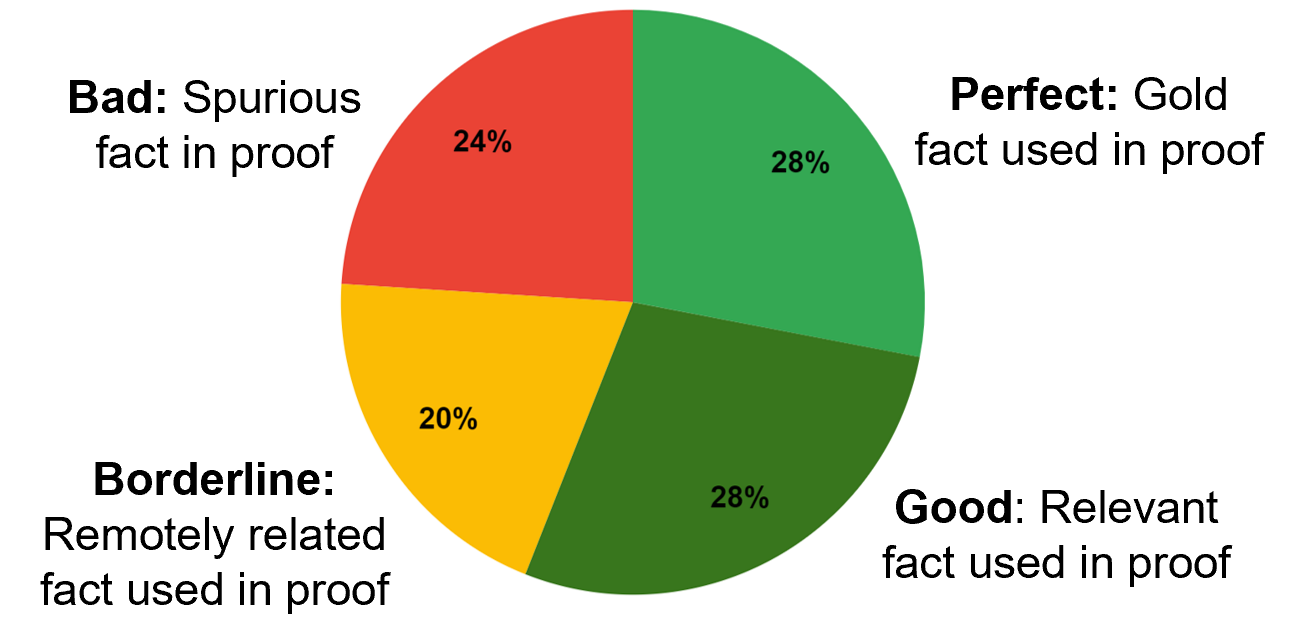}	   
\caption{\TeachMe{} was right for the right reasons in $\approx$75\% of its correct answers. (Examples are shown in Table~E1 in Appendix~E).
\label{successes}}
\vspace{-3mm}
\end{figure} 

\subsubsection{Failure Analysis \label{failure}}

\eat{
In cases where retrieved feedback did not help on new questions, there are several candidate reasons:
\begin{des}
  \item[{\bf Knowledge:}] the relevant knowledge was simply not in memory
  \item[{\bf Retrieval:}] the knowledge was there but not retrieved
  \item[{\bf Reasoning:}] the knowledge was there and retrieved, but \TeachMe{} chose to ignore it
  \item[{\bf Scoring:}] the knowledge was retrieved and used, but the correct answer was still not chosen
  	as a proof for a different answer option scored higher
\end{des}
}

In cases where retrieved feedback did not help on new questions, there are four failure modes:
{\bf knowledge} (the relevant knowledge was simply not in memory); 
{\bf retrieval} (the knowledge was there but not retrieved); 
{\bf reasoning} (the knowledge was there, retrieved, but \TeachMe{} chose to ignore it); 
and {\bf scoring} (the knowledge was retrieved and used, but the proof for a different answer option scored higher).
To measure the relative frequency of these, we examine 50 randomly sampled failure cases,
described below and illustrated in Figure~\ref{failures} (Table~E2 in Appendix~E provides examples), and found: \\
  \textbf{24\% (12/50) missing knowledge:} The gold science fact for the test question was not present in the corpus. Instead, the model tried to make use of the facts retrieved from the corpus to construct proofs but ended up selecting a wrong answer option.  \\
  \textbf{54\% (27/50) bad retrieval:} The gold science fact for the test question was present in the
  corpus but the IR module failed to retrieve it among the top-k.   \\
   \textbf{12\% (6/50) bad reasoning:} The proof generated for the gold answer option was not good, even when the retrieval was good. In 5/6 cases, the model created a bad proof, even though it had correctly started with the correct fact. In the remaining case, the gold core fact was retrieved but then ignored. \\ 
   \textbf{10\% (5/50) bad scoring:} While a good proof for the right answer was generated, it was not scored highest
   either due to some of its (true) premises or entailment being disbelieved by the model, or a false premise or
   bad entailment for a wrong answer being scored highly. Again, further training of the verifiers would help
   alleviate this problem.

\begin{figure}
\centering
     \includegraphics[width=0.9\columnwidth]{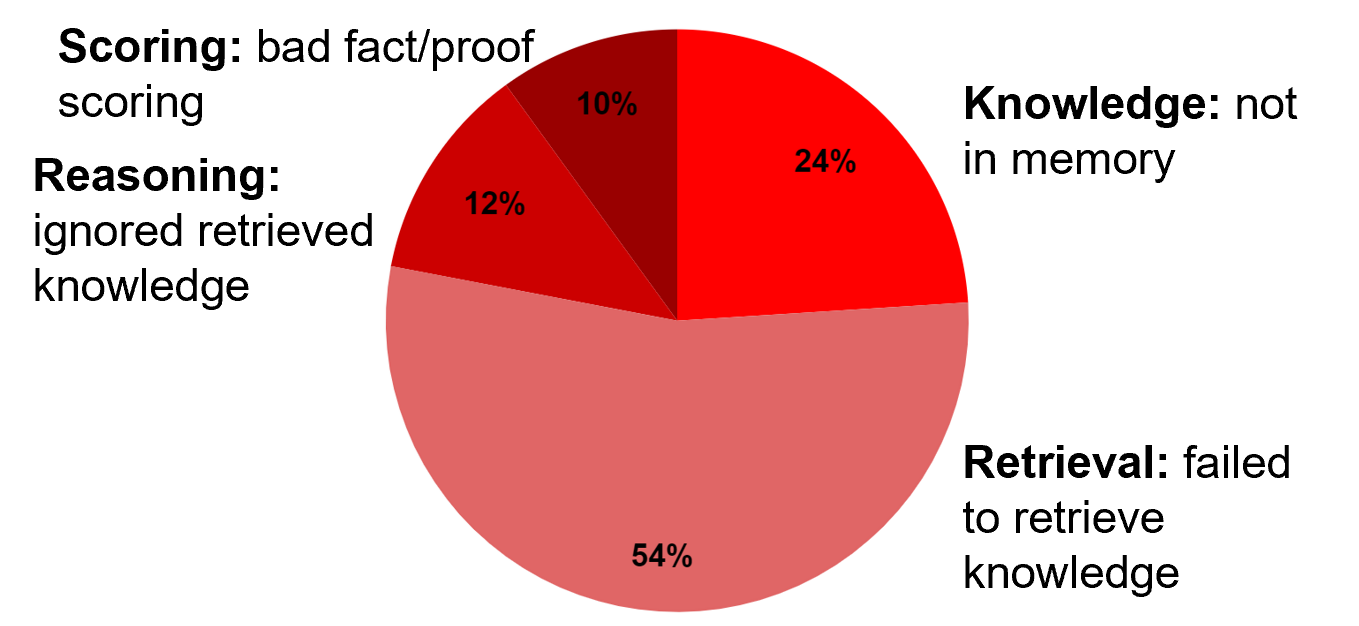}	   
\caption{Causes of failure (\%) for \TeachMe{}'s incorrect answers. (Examples in Table~E2 in Appendix~E).
\label{failures}}
\vspace{-3mm}
\end{figure} 

\subsection{Experiments with Real Users \label{real-users}}

We also ran a small-scale experiment with real users, to test whether users could in practice improve the system's performance.
For this, we took 31 questions from OBQA, based on five core facts, that \TeachMe{} struggled with (getting 20/31 of the questions wrong).
We then split them into a training set containing 1 failing question for each core fact (total 5 questions), and the remaining 26 questions
as a test set. Our interests were (a) whether users could successfully interact with the system to
identify and correct \TeachMe{}'s erroneous beliefs about the 5 training
questions, so it could answer them correctly, and then (b) whether the result of this teaching carried over
to improved performance on the test set. Transcribed examples of some of the dialogs are in Appendix~D.

\begin{figure}
\centering
     \includegraphics[width=0.9\columnwidth]{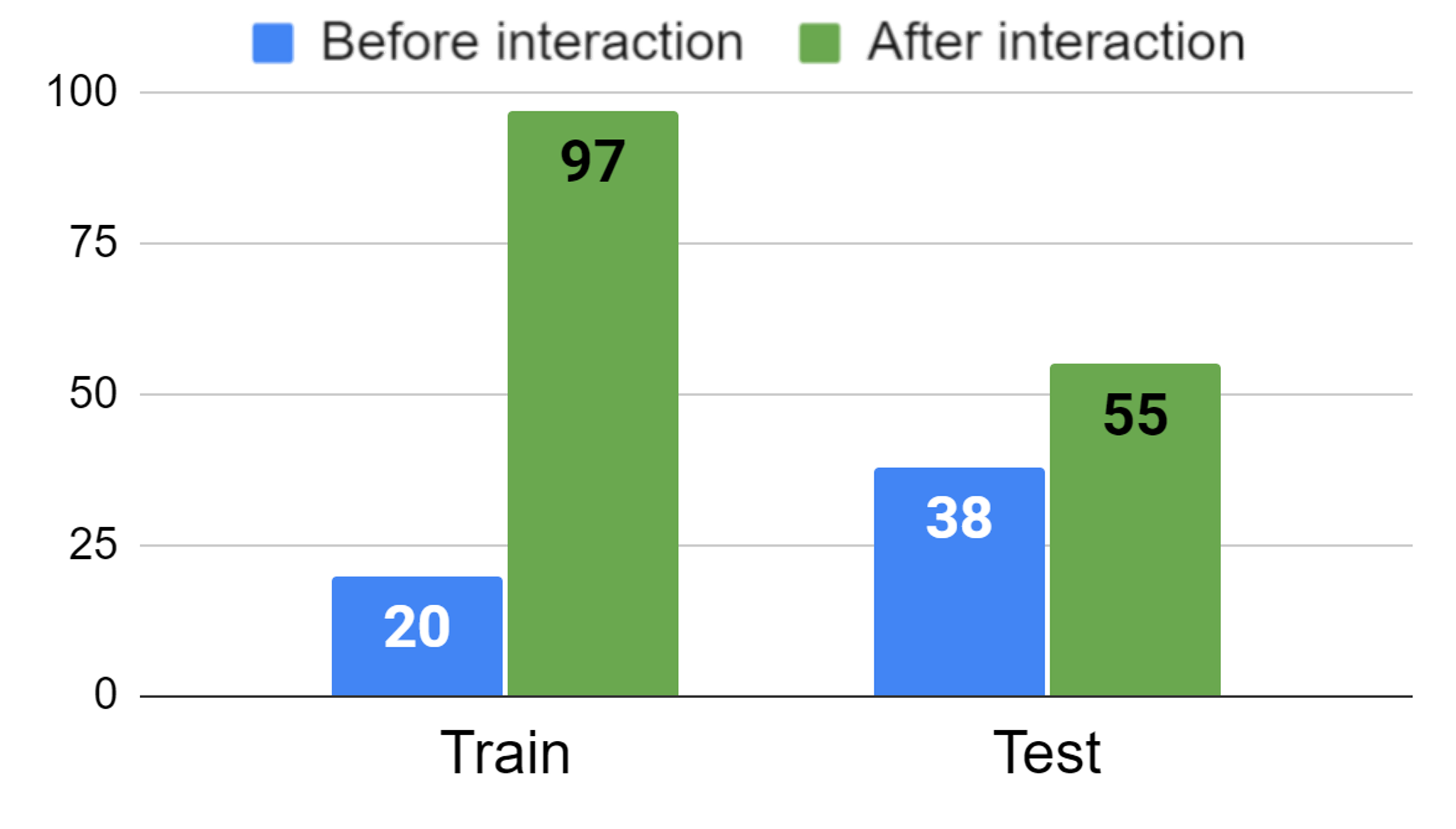}
\vspace{-1mm}     
\caption{\TeachMe{}'s performance (\% correct) substantially improves on a hidden test set (from 38\% to 55\%), a subset of OBQA, 
after users correct/expand its knowledge for the training questions. (Results are averaged over 8 users).
\label{real-user-results}}
\vspace{-3mm}
\end{figure} 

\subsubsection{Results}

The results were averaged over eight users (from within our organization), and are shown in Figure~\ref{real-user-results}
showing \TeachMe{}'s scores before and after user interaction.
On average, users made 2.7 teaching actions per question to correct the system (13.5 per user session for correcting the five questions),
with distribution (\%) as follows (see Appendix~C for details of these categories): fact is missing (24\%), fact is false (12\%), fact is true (6\%), bad reasoning (5\%), fact is irrelevant (5\%), use old fact (10\%), use new fact (37\%).
The average completion time for the task was 19 mins (ranging from 13 to 31 mins). As shown in Figure~\ref{real-user-results}
(first two bars), users were able to correct/expand \TeachMe{}'s knowledge to remove
almost all its errors on the training set (raising \TeachMe{}'s training
score to 97\%). More importantly, the taught system's score on the hidden test set increased by 17\% (38\% to 55\%),
indicating {\bf the knowledge provided by the users generalized to the test set}.

\subsubsection{Analysis}

Of the 208 test answers (26 questions x 8 users), 41 answers changed from incorrect to correct,
and 7 changed from correct to incorrect. Of the 41 that changed to correct (based on an analysis of a subset):
$\approx$70\% a relevant fact was recalled and used in a good proof,
$\approx$10\% the recalled facts altered the model behavior so it generated a good proof with a (generated) relevant fact,
while $\approx$20\% had bad proofs but (fortuitously) scored highest.

\eat{
\begin{ite}
\item $\approx$50\% a relevant fact was recalled and used in a good proof
\item  $\approx$20\% a slightly relevant fact was recalled and used in the good proof
\item  $\approx$10\% the recalled facts altered the model behavior so it generated a good proof with a (generated) relevant fact
\item  $\approx$20\% the newly generated proof was bad but (fortuitously) scored highest
\end{ite}
}
\noindent
For example, for the question:
\quotebox{
{\it Some birds find locations with (A) landmarks (B) road signs (C) eggs (D) \textbf{magnetic patterns} 
} 
}
the model originally selected a wrong answer (eggs), and could not generate a proof for the correct answer.
With memory, its retrieval included the user-supplied fact {\it "Animals can use magnetic patterns to navigate."},
providing crucial knowledge that the model apparently did not know, and allowing a proof for the right answer to be found.

Similarly for the 7 cases that changed from correct to incorrect: about half the time (4/7) the system
did recall a relevant fact, but either ignored it (2/7) or generated a bad proof (2/7). In the remaining
3/7 cases, there was no relevant fact retrieved, but the retrievals served to confuse the generator.
For example, for the question: 
\quotebox{
{\it Gills are used to breath water by what? (A) \textbf{salmon} (B) fishing boats (C) penguins...} 
}
the system originally selected the right answer (salmon), with an (incorrect) proof for penguins close behind.
With memory, it retrieved the user-supplied fact {\it "Animals can use magnetic patterns to navigate."},
irrelevant to the question, but enough when added to the context to slightly change the verification scores,
resulting in the (bad) proof for penguin being scored highest.

\section{Discussion and Conclusion \label{Conclusion}}

Our goal is a teachable reasoning system, where users can interact to see
its beliefs and reasoning, and correct it when it is wrong.
We have shown that by embedding an entailment-based QA model in a larger system with a dynamic,
persistent memory, users can correct and override model beliefs, resulting in an
overall system that can improve over time without retraining. To our
knowledge, this is the first system to show that user-provided and model-internal
beliefs can be integrated together for systematic reasoning.
This is significant as it is a step towards systems that can not only interact with users, but continually
learn from them.

Although we have created and evaluated an integrated system, 
numerous issues still remain. For reasoning, methods to avoid uninteresting
(near-tautologous) proofs are needed. For interaction, we have treated
``teaching'' primarily as question-centric debugging, but clearly there are other
styles to explore.
Finally while the memory usefully biases \TeachMe{} for new tasks, the
effects of placing new knowledge in an input context are not fully predictable,
despite careful training. These are all areas for future exploration.

Despite these, the research agenda is an exciting one, pointing towards
future systems that can learn directly from users in a conversational
way, rather than solely training on large datasets. It also suggests
a way of overcoming the opaqueness of neural systems, by
viewing models as components in a larger system with a persistent memory
and that can systematically reason. We look forward to future developments
in these directions.



\section*{Limitations}

We have shown how a dynamic memory, paired with a QA system that can provide faithful
explanations, can allow users to correct erroneous system beliefs, and thus improve its
performance without model retraining. While exciting, there are several limitations
with the current approach and opportunities for future work.

First, we have so far only worked with relatively small memories
(up to $\approx$2000 facts, for the simulated users, Section~\ref{simulation}). A deployed system could
potentially acquire orders of magnitude more user-supplied facts, raising challenges for
retrieval and memory management. Eventually, one might want to retrain the model
to incorporate these new/corrected beliefs into the model itself.

Second, as memory grows, it is possible that conflicting facts may arise in it,
either from a user being inconsistent, or assuming different contexts for a fact,
or from different users. Mechanisms for belief management would be advantageous
to spot and repair such problems, e.g., \cite{beliefbank}.

Third, the approach relies on the system generating meaningful chains of reasoning for
its answers (in particular, for its incorrect answers) to engage the user.
However, in some cases those chains are poor (Section~\ref{failure}), 
and could be improved through enhanced proof generation techniques.

In addition, two broader themes merit more exploration. First, we have
treated ``teaching'' as question-centric debugging, but clearly there are broader
styles to explore, e.g., the user volunteering general knowledge
up-front, probing what the system already knows, and following a curriculum.
Second, we have assumed a single-user environment dealing with factual questions,
but a deployed system may encounter users with different beliefs about the world,
and/or different opinions. This problem is not new and mechanisms exist
to handle this (e.g., for Wikipedia), but would need to be integrated into
this environment too for large-scale deployment.

\camera{
Finally, our approach relies on human feedback  on new questions that \teachme{} fails to answer or fails to justify indicating significant human efforts. We are exploring three mechanisms for reducing such human efforts: (a) \teachme{} can spot some errors itself by using external text sources to verify them (b) \teachme{} can carefully order the teaching questions. That way, if the user can debug some critical system misconceptions early, then many future questions will be answered correctly (hence not requiring user input).  (c) Ask multiple users, e.g., factoring the teaching task into a curriculum of smaller topics ( ``magnetism'', ``gravity'', ``adaptation'' etc.) for different users to work on.}

\eat{
  Paper 2
 - Bad proofs slip through the entailment verifier
 - Not explored comprehension-based questions: While C can accomodate this, we haven't tried this
 - Subjectivity and handling different opinions
 - Only explored small memory - with a large memory, issues of recall and speed become an issue
 - Effect of contextual feedback helps but not always in a predictable way
 - Simplistic teaching - real teaching

Future work:
 - Better proof quality
 - Explanations as deduction is a narrow 
 - Better notion of teaching - encouraging a systematic portrail of a domain, rather than debugging
 - Memory management for large KBs
 - Consistency maintenance
 - Working with multiple opinions
 - Role of context
 - Predictable impact of context

Future work:
 - Better proof quality
 - Consistency maintenance
 - Predictable impact of context

 - Memory management for large KBs
 - Better notion of teaching - encouraging a systematic portrail of a domain, rather than debugging
 - Working with multiple opinions
}

\section*{Acknowledgements}
This research was made possible, in part, by funding from Open Philanthropy.
We also thank Google for providing the TPUs for conducting experiments.
Finally thanks to the end-users and annotators and who participated in experiments and
data collection, and to Ashish Sabharwal for valuable feedback on an earlier draft.



\bibliography{references}
\bibliographystyle{acl_natbib}


\clearpage

\appendix



\section*{Appendix A. Different Memory Indexing Strategies}

As described in Section \ref{retrieval}, \TeachMe{} retrieves up to r (=5) sentences from memory using the question as the search query, using a standard BM25 search algorithm. We evaluate the following alternative ways of indexing this memory (in all 4 ways, the ``document'' is always the fact collected through interaction but the indexing terms are different): \\
\textbf{1) F:} Index by the terms in the fact \\
\textbf{2) Q:} Index by the terms in the question for which the fact was provided as feedback \\
\textbf{3) Q + F:} Index by concatenation of the question and associated fact
(For both options 2, 3, if a fact is useful for multiple questions, it will appear multiple times in the memory.)\\
\textbf{4) Relevant Qs + F:} Index by concatenation of the fact and all the questions it is relevant to (Each fact appears only once in the memory.)

Table~A1 compares the retrieval performance of these four indexing strategies on OBQA Dev questions, using the simulated user setup, where we measure how well the gold fact associated with a Dev question is retrieved, using an index built from the Train questions (that also use these gold facts). 
In all cases the search query is the question. 
We find the simple strategy of indexing by the fact itself performs the best (used in rest of the experiments in this paper).

\begin{table}[h]
\centering
{\small
\begin{tabular}{l|rrrrr} \hline
Index by  & R@1   &	    R@2	 &   R@3 &	    R@5	&    R@10 \\ \hline
F	& 31.0	& 39.2  &	44.2 &	51.0  &	58.8 \\
Q	& 19.0	&27.2	&30.8	&36.6	&44.0\\
Q + F	& 22.0	&29.4	&35.8	&41.2	&50.2\\
Relevant Qs + F &	10.6	&14.0	&17.0	&19.4	&25.6\\
\hline
\end{tabular}
}
\caption*{Table A1: Recall of gold fact for OBQA Dev questions when \TeachMe{} indexed the gold facts for Train questions in four different ways.}
\end{table}


\twocolumn[\section*{Appendix B. \TeachMe{}'s Interactive Interface} \label{appendix:interactive_interface}]


\vspace{2mm}

 \fbox{\includegraphics[width=0.8\textwidth]{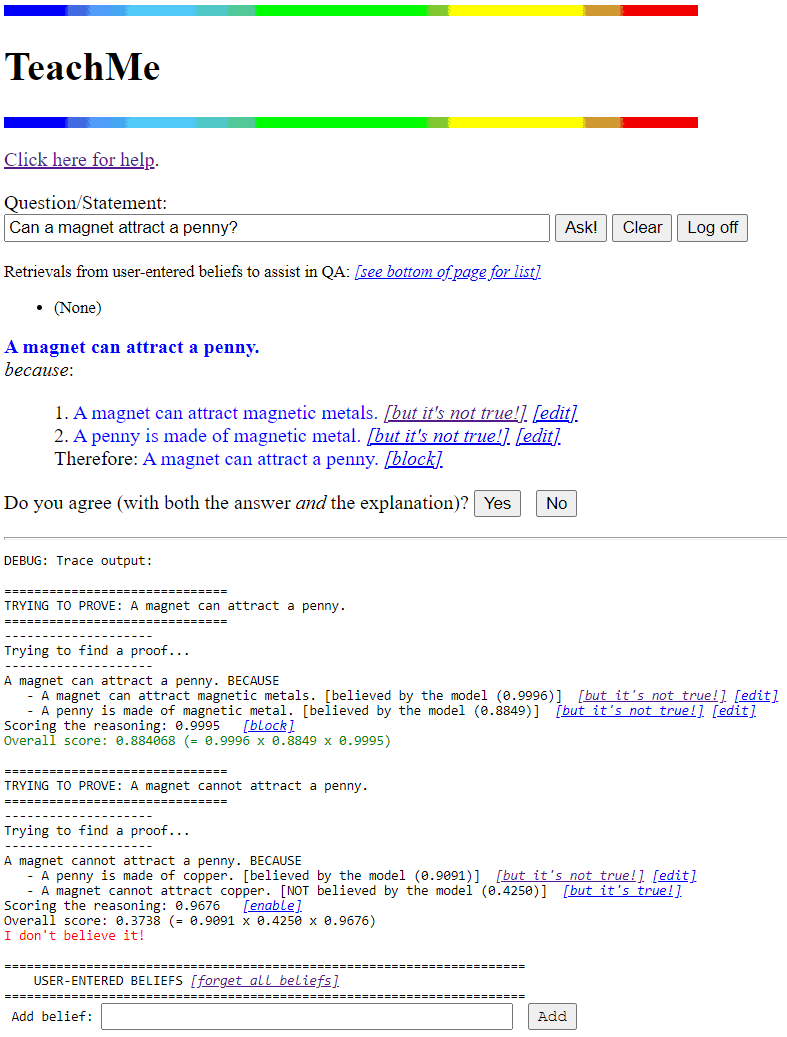}}	   


\clearpage





\section*{Appendix C: Example of Graphical Interaction}

Below shows snippets of the graphical interface with TeachMe,
walking through a path of interaction in the dialog tree shown in Figure~\ref{fig:interaction_flow}.
First, the user (playing the role of teacher) has asked \TeachMe{}
the question ``Can a magnet attract a penny?'' in the upper box.
(The correct answer is ``no'', as pennies are made of the non-magnetic metal copper). The system
has responded with an (incorrect) answer and proof (blue):

\noindent \fbox{\includegraphics[width=1\columnwidth]{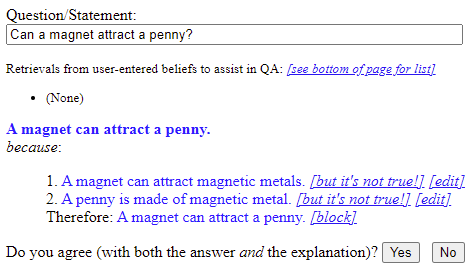}}	   

\noindent
The error in this case is the system's belief that ``A penny is made of magnetic
metal.'' (pennies are in fact made of non-magnetic copper). To indicate the error,
the user clicks on the ``No'' button in response to ``Do you agree?'', and then here chooses to correct
the error by entering the missing, required knowledge that ``A penny is made of copper.'':

\noindent \fbox{\includegraphics[width=1\columnwidth]{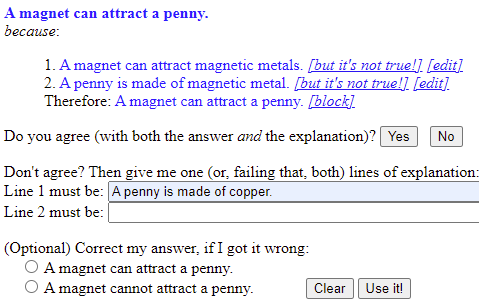}}	   

\noindent
On reasking the question, \TeachMe{} uses this user-supplied fact
as the first part of the proof (via forced generation), 
but still gets the answer wrong due to a different misunderstanding, namely that
``A magnet can attract copper.''

\noindent \fbox{\includegraphics[width=1\columnwidth]{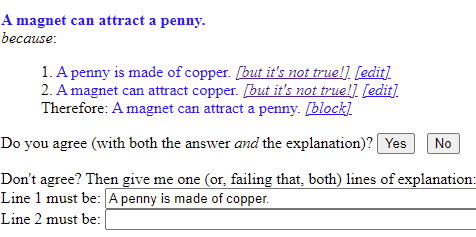}}	   

\noindent
Here the user chooses to correct this mistake by clicking on
{\it \blue{[but it's not true!]}} next to ``A magnet can attract copper'', causing the opposite fact
(``A magnet cannot attract copper'') to be added to memory.
Finally the system gets the answer right:

\noindent \fbox{\includegraphics[width=1\columnwidth]{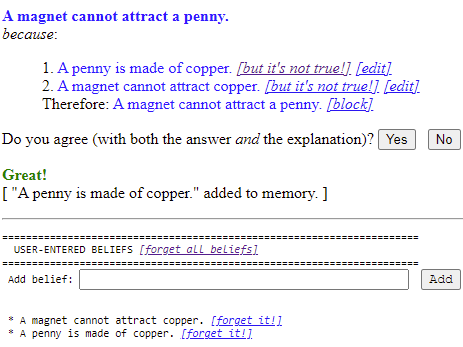}}	   

Confirming the answer causes \TeachMe{} to note both the
facts, conclusion, and proof in memory, so that it will produce
the same answer if the question is reasked. In addition, the stored
facts can be retrieved to help answer future questions, a key role
for the memory.

\clearpage

\section*{Appendix D: Additional Examples of Interactions}

We here provide some additional examples of real user dialogs with the system, transcribed into textual format
(the original interactions are via the graphical interface). These examples are all from the
user trials in Section~\ref{real-users}, using questions from the OBQA dataset, and are all cases
where \TeachMe{} initially answered incorrectly.
The dialogs illustrate how the user (blue) used different actions (annotated in bold
square brackets, referencing actions in Figure~C1, Appendix C) to diagnose and correct the system's misunderstandings.

\ \\
\small{
\begin{des}
\item[{\bf U:}] \blue{Gills allow (A) flounder to intake oxygen (B) giraffes to breathe underwater (C) respiration for sea going birds}
\item[{\bf S:}] Gills allow respiration for sea going birds \bfit{because:}: \\
   1. Gills are used for breathing by aquatic animals \\
   2. A sea going bird is a kind of aquatic animal \\
   Do you agree? 
\item[{\bf U:}] \blue{No. Birds do not have gills.}		\hfill		\bfit{[A fact is missing]}
\item[{\bf S:}] Gills allow flounder to intake oxygen \bfit{because:} \\
   1. Gills are used for breathing by aquatic animals \\
   2. A flounder is a kind of aquatic animal \\
   Do you agree? 
\item[{\bf U:}] \blue{yes}					\hfill		\bfit{[Looks good]}
\vspace*{2mm} \hrule \vspace*{2mm}
\item[{\bf U:}] \blue{An example of a fossil is a paw print in what? (A) water (B) sky (C) air (D) hard stones}
\item[{\bf S:}] I can't find an answer! \\
   Here's the facts that I considered: \\
   1. Sedimentary rocks are made of hard stones. [\bfit{but} I think this is false!] \\
   2. A paw print is a kind of imprint. \\
   ... \\
   6. Sedimentary rock is made of air. [\bfit{but} I think this is false!] \\
   7. An example of a fossil is a paw print in sedimentary rock. [\bfit{but} I think this is false!] \\
   Which fact should I use? 
\item[{\bf U:}] \blue{7 is true.}				\hfill		\bfit{[A disbelieved fact is true]}
\item[{\bf S:}] I can't find an answer! \\
   Here's the facts that I considered: \\
   1. Sedimentary rocks are made of hard stones. [\bfit{but} I think this is false!] \\
   2. A paw print is a kind of imprint. \\
   ... \\
   6. Sedimentary rock is made of air. [\bfit{but} I think this is false!] \\
   7. An example of a fossil is a paw print in sedimentary rock. \\
   Which fact should I use? 
\item[{\bf U:}] \blue{use 1.}				\hfill		\bfit{[Use one of these facts]}
\item[{\bf S:}] An example of a fossil is a paw print in hard stones. \bfit{because:} \\
   1. An example of a fossil is a paw print in sedimentary rock. \\
   2. Sedimentary rock is made of hard stones. \\
   Do you agree? 
\item[{\bf U:}] \blue{yes}				\hfill			\bfit{[Looks good]}

\vspace*{2mm} \hrule \vspace*{2mm}
\item[{\bf U:}] \blue{Clouds flit and flow (A) in a bank (B) on the internet (C) above our heads (D) in cloudless skies}
\item[{\bf S:}] I can't find an answer! \\
   Here's the facts that I considered: \\
   1. Cloudless skies do not have clouds. \\
   2. Clouds move in the sky. \\
   3. The internet does not have clouds. \\
   4. A bank is a kind of formation. \\
   Which fact should I use? 
\item[{\bf U:}] \blue{use 2}				\hfill		\bfit{[Use one of these facts]}
\item[{\bf S:}] Clouds flit and flow above our heads. \bfit{because:} \\
   1. Clouds move in the sky. \\
   2. Clouds flit and flow. \\
   Do you agree? 
\item[{\bf U:}] \blue{No. use The sky is above our heads.}	\hfill		\bfit{[Use a different fact]}
\item[{\bf S:}] Clouds flit and flow above our heads. \bfit{because:} \\
   1. Clouds move in the sky. \\
   2. The sky is above our heads. \\
   Do you agree? 
\item[{\bf U:}] \blue{yes}					\hfill		\bfit{[Looks good]} 
\vspace*{2mm} \hrule \vspace*{2mm}
\item[{\bf U:}] \blue{What's used by migrating animals to find locations? (A) GPS satellites (B) our planet's magnetic patterns ...} 
\item[{\bf S:}] Migrating animals use GPS satellites to find locations \bfit{because:} \\
   1. Migrating animals need to find locations. \\
   2. GPS satellites can be used to find locations. \\
   Do you agree?
\item[{\bf U:}] \blue{No. Bad reasoning.}		\hfill			\bfit{[Facts ok, but reasoning is bad]}   
\item[{\bf S:}] I can't find an answer! \\
   Here's the facts that I considered: \\
   1. Our planet's magnetic patterns are formed by the Earth's magnetic field. \\
   ... \\
   7. GPS satellites can be used to find locations. \\
   Which fact should I use? 
\item[{\bf U:}] \blue{Migrating animals have an internal magnetic} \\
\blue{compass.}  \hfill  \bfit{[A fact is missing]} 
\item[{\bf S:}] Migrating animals use our planet's magnetic patterns to find locations \bfit{because:} \\
   1. Migrating animals have an internal magnetic compass. \\
   2. Migrating animals use compass to find locations. \\
   Do you agree? 
\item[{\bf U:}] \blue{yes}			\hfill				\bfit{[Looks good]} \\
\hrule
\end{des}
}

\begin{table*}[h!]
  {\large \bf Appendix E: Success and Failure Analysis: Examples \\
    \    \\
    \ }
\centering
{\small
\setlength{\tabcolsep}{2pt}	
\begin{tabular}{llll} \hline
{\bf Category} & {\bf Question }  & \makecell[l]{\bf Model generated \\ \bf answer, best proof } & \textbf{Comments}\\
\hline
 \makecell[l]{Uses Gold \\ core fact} & 
 \makecell[l]{\\ Q: Owls are likely \\ to hunt at \\ (A) 3pm \textbf{(B) 2am} \\(C) 6pm (D) 7am
} & 
\makecell[l]{
Owls are likely to hunt at \green{2am} {\bf because:}\\
\sgap Nocturnal predators hunt during the night. \\ \sgap Owls are a kind of nocturnal predator. \\ \sgap 2:00 AM is after midnight. 
}  &
\makecell[l]{ 
Model generated reasonable \\
chain of reasoning justifying\\
the correct answer \\
using the gold core fact. 
} \\
\hline
 \makecell[l]{Uses relevant\\ core fact} & 
 \makecell[l]{\\ Wax can be used \\ similarly to\\
 (A) wood \textbf{(B) rubber} \\
 (C) water (D) metal
} & 
\makecell[l]{
Wax ...similarly to \green{rubber} {\bf because:} \\
\sgap Rubber is a thermal insulator.\\
\sgap Wax is a thermal insulator.
}  &
\makecell[l]{ 
Gold core fact: ``wax is a electrical \\ 
insulator''  was not in top-k retrieval. \\
But the model used another relevant \\ 
fact to construct a good proof.
} \\
\hline
 \makecell[l]{Uses remotely\\ relevant fact} & 
 \makecell[l]{Q: They looked where \\ the log decayed  to garden \\ as it would leave the earth\\ \textbf{(A) richer} (B) dryer \\(C) sandy (D) harder
} & 
\makecell[l]{
...leave the earth \green{richer} {\bf because:}\\
\orange{\sgap The mantle is a layer of the Earth.} \\
\sgap Decaying leaves a layer of soil \\ \ \ \ \ \ \ \ on the Earth. \\
\sgap Soil is a kind of resource. \\
\sgap If something decays then that  something  \\ \ \ \ \ \ \ \ will leave the earth richer.
}  &
\makecell[l]{ 
  Model includes a remotely relevant \\
  fact in the proof. As the answer\\
  is still entailed, the verifier \\
  approves the proof.} \\
\hline
 \makecell[l]{Uses spurious \\ facts} & 
 \makecell[l]{Q: Dunes can be made  \\ out of the same thing as \\ product of a rain storm? \\ (A) clothes (B) food \\ (C) forests \textbf{(D) castles}
} & 
\makecell[l]{
Dunes...same thing as...\green{castles} {\bf because:} \\
\sgap \red{Some birds live in forests.} \\
\sgap \red{Some rabbits live in forests.} \\
\sgap Some dunes can be made out of \\ \ \ \ \ \ \ \ the same thing as castles.
}  &
\makecell[l]{
  The model includes irrelevant facts \\
  in the proof. As the answer is \\
  still entailed, the verifier does not \\
  catch the redundancy.} \\
\hline
\end{tabular}
}
\caption*{Table E1: Examples of success cases where \TeachMe{} answered a test question accurately by justifying the gold answer choice
  (see Section~\ref{success} and Figure~\ref{successes}).
  \label{table:success_examples}}
\vspace{-2mm}
\end{table*}

\begin{table*}[h!]
\centering
{\small
\setlength{\tabcolsep}{2pt}	
\begin{tabular}{llll} \hline
{\bf Category} & {\bf Question }  & \makecell[l]{\bf Model generated answer,\\ \bf best (here, bad) proof } & \textbf{Comments}\\
\hline
 \makecell[l]{Missing \\ knowledge} & 
 \makecell[l]{\\ Q: An ice cube placed in \\sunlight will \\ \textbf{(A) shrink} (B) change color \\(C) grow (D) freeze
} & 
\makecell[l]{
An ice cube ... will \red{grow} {\bf because:}\\
\sgap A glacier is made of ice. \\
\sgap A plant requires sunlight  \\ \ \ \ \ \ \ \ to grow. 
}  &
\makecell[l]{ 
Gold science fact: \\ "Sunlight produces heat" \\ 
is not part of memory  
} \\
\hline
 \makecell[l]{Bad \\ IR} & 
 \makecell[l]{\\ Q: Which of the following \\ is warm blooded? \\ (A) toad (B) snake \\ (C) turtle \textbf{(D) skunk}
} & 
\makecell[l]{
A \red{toad} is warm blooded {\bf because:}\\
\sgap A warm something is a source  \\ \ \ \ \ \ \ \ of heat. \\ \sgap A toad is a kind of animal. \\ \sgap Animals are warm blooded.
}  &
\makecell[l]{ 
Gold science fact: "A mammal is\\ warm-blooded." 
is part of memory  \\ but not in top-k retrieval
} \\
\hline
 \makecell[l]{Bad \\ Reasoning} & 
 \makecell[l]{Q: What has more gravity \\ force than Earth but \\ less than the sun? \\ \textbf{(A) Jupiter} (B) the moon \\(C) a space station (D) a comet
} & 
\makecell[l]{
\red{The moon} has ... {\bf because:} \\
\sgap The Moon orbits the Earth. \\ \sgap The Earth has more mass than \\ \ \ \ \ \ \ \ the Moon. \\ \sgap The Moon has more gravity  \\ \ \ \ \ \ \ \ force than the Earth. \\ \sgap The Sun has more mass than  \\ \ \ \ \ \ \ \ the Moon.
}  &
\makecell[l]{ 
Gold science fact: "as mass of a \\ celestial body increases, \\ the force of gravity on that planet \\ will increase" \\ 
is at rank 4 in the retrieval. \\
Model incorrectly starts with \\ less relevant fact and \\completes a proof for the wrong \\answer option.
} \\
\hline
 \makecell[l]{Bad \\ Scoring} & 
 \makecell[l]{Q: Which of these is required \\ for a plant to enjoy the \\ product of a rain storm? \\ \textbf{(A) xylem} (B) luck \\ (C) magic (D) dirt
} & 
\makecell[l]{
\red{Dirt} is required...{\bf because:}\\
\sgap Clouds produce rain. \\ \sgap A plant requires dirt to grow.
}  &
\makecell[l]{ 
Model generated valid proof for correct \\ answer ``xylem'' using Gold science fact \\ (from top-k retrievals): "Xylem carries water \\from  the roots of a plant  to the leaves of a \\plant." but premise verifier scored it as \\
incorrect premise (score=0.045). Hence the \\
proof for wrong answer was scored higher \\
than that for the correct answer.
} \\
\hline
\end{tabular}
}
\caption*{Table E2: Examples of four different failure categories where \TeachMe{} answered a test question incorrectly, providing a bad proof for the wrong answer (the correct answer option is boldfaced).
  See Section~\ref{failure} and Figure~\ref{failures}. \label{table:failure_examples}}
\end{table*}

\end{document}